\documentclass[10pt,twocolumn,letterpaper]{article}

\usepackage[pagenumbers]{cvpr} 

\usepackage{multirow}

\newcommand{\gd}[1]{\textcolor{red}{#1}}

\definecolor{cvprblue}{rgb}{0.21,0.49,0.74}
\usepackage[pagebackref,breaklinks,colorlinks,allcolors=cvprblue]{hyperref}
\usepackage{amsmath}
\usepackage{amssymb}
\usepackage{booktabs}
\usepackage{multirow} 
\usepackage{float}

\def\confName{CVPR}
\def\confYear{2026}

\title{ADNet: A Large-Scale and Extensible Multi-Domain Benchmark for Anomaly Detection Across 380 Real-World Categories
\vspace{-1em}
}

\author{%
    Hai Ling$^{1,2}$ \hspace{1.5em} Jia Guo$^{3}$ \hspace{1.5em} Zhulin Tao$^{1*}$ \hspace{1.5em} Yunkang Cao$^{4}$ \hspace{1.5em} Donglin Di$^{3}$ \\ 
    Hongyan Xu$^{5}$ \hspace{1.5em} Xiu Su$^{5}$ \hspace{1.5em} Yang Song$^{6}$ \hspace{1.5em} Lei Fan$^{2,6\dagger}$\thanks{Corresponding Authors: \texttt{lei.fan1@unsw.edu.au}, \texttt{taozl@cuc.edu.cn} \\ \mbox{}\hspace{1.1em} $^{\dagger}$Project Leader.} \\
    \vspace{-1em}
    \\
    $^{1}$Communication University of China \hspace{1.5em} 
    $^{2}$DZ Matrix \hspace{1.5em} 
    $^{3}$Tsinghua University \\
    $^{4}$Hunan University \hspace{1.5em} 
    $^{5}$Central South University \hspace{1.5em} 
    $^{6}$UNSW Sydney
    \vspace{-1.8em}
}

\begin{document}
\maketitle
\begin{abstract}
\vspace{-20pt}

Anomaly detection (AD) aims to identify defects using normal-only training data. Existing anomaly detection benchmarks (e.g., MVTec-AD with 15 categories) cover only a narrow range of categories, limiting the evaluation of cross-context generalization and scalability. 
We introduce \textbf{ADNet}, a large-scale, multi-domain benchmark comprising \textbf{380} categories aggregated from 49 publicly available datasets across Electronics, Industry, Agrifood, Infrastructure, and Medical domains. The benchmark includes a total of \textbf{196,294} RGB images, consisting of 116,192 normal samples for training and 80,102 test images, of which 60,311 are anomalous. All images are standardized with MVTec-style pixel-level annotations and structured text descriptions spanning both spatial and visual attributes, enabling multimodal anomaly detection tasks. 
Extensive experiments reveal a clear scalability challenge: existing state-of-the-art methods achieve $90.6\%$ I-AUROC in one-for-one settings but drop to $78.5\%$ when scaling to all 380 categories in a multi-class setting.
To this end, we propose \textbf{Dinomaly$^m$}, a context-guided Mixture-of-Experts extension of Dinomaly that expands decoder capacity without added inference cost. It achieves $83.2\%$ I-AUROC and $93.1\%$ P-AUROC, demonstrating superior performance over existing approaches.
We aim to make ADNet a standardized and extensible benchmark, supporting the community in expanding anomaly detection datasets across diverse domains and providing a scalable foundation for AD foundation models. Dataset: \url{https://grainnet.github.io/ADNet}.

\end{abstract}
\vspace{-15pt}
\section{Introduction}
\label{sec:intro}

Unsupervised anomaly detection (AD) in visual data aims to identify anomalous regions or defective samples using only anomaly-free samples for training. This topic is critical in safety-sensitive and high-reliability scenarios such as industrial inspection, medical imaging, and infrastructure monitoring. Its development has been largely driven by established benchmarks such as MVTec-AD~\cite{mvtec}, VisA~\cite{visa}, and BTAD~\cite{btad}, which provide diverse object categories. These datasets not only enable fair comparison but also drive the exploration of different modeling paradigms, such as reconstruction-based methods~\cite{deng2022anomaly,fan2024revitalizing}, data synthesis-based methods~\cite{li2021cutpaste,zhang2024realnet,fan2024patch}, embedding-based methods~\cite{liang2025tocoad}, and memory-based methods~\cite{patchcore}.

\begin{figure}[t]
\centering
\includegraphics[width=\columnwidth]{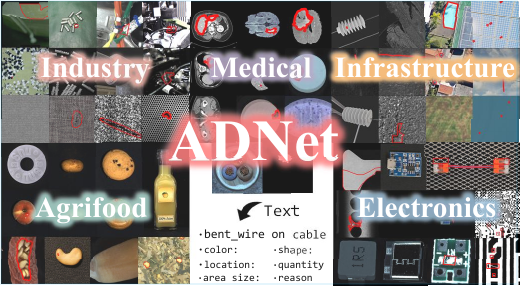}
\caption{Overview of ADNet, covering 380 categories across five application domains.}
\label{fig:adnet}
\vspace{-16pt}
\end{figure}

\begin{figure*}[t]
\centering
\includegraphics[width=\textwidth]{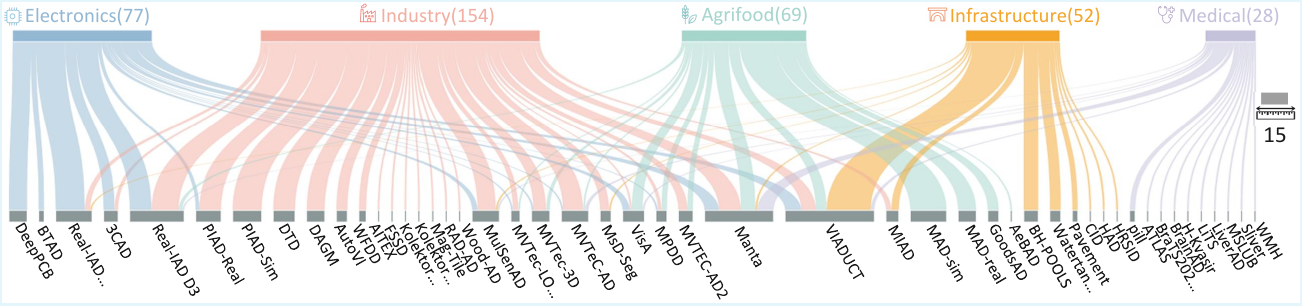}
\caption{\textbf{ADNet is constructed by consolidating 49 publicly available AD datasets}. We validate their composition and quality, and hierarchically integrate all sources into five target domains: Electronics, Industry, Agrifood, Infrastructure, and Medical.}
\vspace{-10pt}
\label{fig:dataset_overview}
\end{figure*}

Early AD methods focus on training individual models for each category, commonly referred to as the 'one-for-one' setting. However, maintaining separate models for numerous categories is resource-intensive and difficult to manage during deployment. To overcome this limitation, recent efforts~\cite{uniad,dinomaly} have shifted toward 'one-for-all' approaches that aim to build a unified model capable of handling multiple categories simultaneously. Both settings have achieved remarkable performance, consistently attaining near-perfect I-AUROC scores ($> 95\%$) on standard benchmarks.
These results highlight the rapid progress in model design but also reveal the inherent limitations of existing datasets.
Although recent datasets extend AD to broader domains such as consumer electronics~\cite{3CAD,DeepPCB}, medical imaging~\cite{BraTS2021,Brain_AD}, tiny objects~\cite{manta}, and new imaging modalities~\cite{mulsenad,HAD}, they still contain a limited number of categories (fewer than 40), focus on a single domain, and are collected under controlled imaging conditions (\textit{e.g.,} fixed lighting and viewpoints).
As a result, current evaluations remain constrained to isolated, small-scale benchmarks. This mismatch raises fundamental questions: \textit{How well do existing AD methods perform when scaled to a large number of categories? And can we build an AD benchmark similar to ImageNet~\cite{imagenet}?}

In this paper, we introduce ADNet (see Figure~\ref{fig:adnet}), a large-scale anomaly detection dataset comprising 380 categories, which is $25\times$ larger in the number of categories than MVTec-AD. 
ADNet consolidates over 49 publicly available AD datasets into a unified benchmark (see Figure~\ref{fig:dataset_overview}), producing 116,192 training and 80,102 test images (including 60,311 anomalous samples). The categories are grouped into five domains: Electronics, Industry, Agrifood, Infrastructure, and Medical. All heterogeneous data sources are standardized into a consistent format with MVTec-style pixel-level annotations, along with expert-provided textual descriptions for each anomalous sample. 

Compared to previous datasets, ADNet enables systematic evaluation of two critical capabilities:
\textbf{\textit{i) Scalability vs. number of categories}.} We evaluate existing methods. While models achieve strong performance in the one-for-one setting (90.6\% I-AUROC), performance drops to 78.5\% in the multi-class setting. This degradation reveals that current models struggle to maintain robustness and scalability as the number of categories grows, limiting their applicability in real-world environments.
\textbf{\textit{ii) Anomaly confusion.}}
ADNet includes categories spanning diverse domains where identical visual patterns can represent opposite anomaly semantics.
For instance, cracks are normal in cookies but defective in printed circuit boards.
Models (\textit{e.g.,} PatchCore~\cite{patchcore}), which rely on memorizing category-specific appearance priors, are prone to confusion in such cases. This challenges models to move beyond superficial pattern matching.

Our contributions are summarized as follows:
\begin{itemize}
    \item We introduce \textbf{ADNet}, a large-scale AD dataset containing 196,294 images across 380 categories spanning five domains.
    \item We provide a unified MVTec-style annotation format, and augment each anomalous sample with fine-grained textual descriptions.
    
    \item We propose \textbf{Dinomaly$^m$}, a strong baseline that integrates a Mixture-of-Experts into Dinomaly~\cite{dinomaly}, improving the I-AUROC from $78.5\%$ to $83.2\%$ on ADNet.
    
\end{itemize}

\section{Related Work}
\label{sec:related}

\subsection{Anomaly Detection Datasets}

Early studies focused on specific application domains such as industrial manufacturing~\cite{mvtec,visa,btad}, medical imaging~\cite{BraTS2021,Hyper-Kvasir,ATLAS}, infrastructure inspection~\cite{CID,Pavement-Defect-Datasets-crack500}, and hyperspectral or agricultural imaging~\cite{HAD,
wood-anomaly-detection-one-class-classification,fan2025grainbrain}. More recent works have expanded along several key directions: 
i) \textit{Dat
a scale and class diversity}. Real-IAD~\cite{realiad} and Kaputt~\cite{kaputt} increased both dataset size and category diversity, with Kaputt demonstrating the feasibility of large-scale collection in retail logistics;
ii) \textit{Multimodal sensing}. Datasets (\eg, MVTec 3D-AD~\cite{mvtec3d}, MulSen-AD~\cite{mulsenad}, Real-IAD D$^3$~\cite{real-iad-d3}) incorporate depth maps, multi-view imagery, and multi-sensor inputs to enhance robustness beyond RGB data; 
iii) \textit{Language-grounded annotation}. MANTA~\cite{manta} provides textual descriptions for anomalies across 38 categories, enabling text-guided anomaly localization; 
and iv) \textit{Specialized scenarios}. MVTec LOCO~\cite{mvtec_loco} focuses on logical rather than visual inconsistencies, while FSSD~\cite{FSSD-12} and MSD~\cite{MSD-seg} explore few-shot settings. Additionally, domain-specific datasets have been developed for textiles~\cite{AITEX}, electronics~\cite{DeepPCB}, and steel~\cite {KolektorSDD}.

Compared to existing datasets, which are domain-specific, limited in category diversity, and vary in data formats and annotations, we introduce ADNet, which unifies the data structure and annotation protocol across over 49 datasets, aggregating 380 categories spanning five domains.

\subsection{Anomaly Detection Methods}
Existing AD approaches have shifted from one-for-one models to more general frameworks to support scalable deployment across diverse scenarios.

\textbf{One-for-One.} This paradigm trains an individual model for each category. Existing methods may be grouped into: i) \textit{Reconstruction-based methods}, which detect anomalies through reconstruction errors from autoencoders~\cite{deng2022anomaly,fan2024revitalizing} or diffusion models~\cite{liu2023diversity,ristea2022self}; ii) \textit{Synthesis-based methods}, which generate pseudo anomalies to enable supervised training~\cite{li2021cutpaste,zavrtanik2021draem,liu2023simplenet,zhang2024realnet,fan2024patch}; iii) \textit{Embedding-based methods}, which learn compact representations of normal data to distinguish anomalies~\cite{ruff2018deep,defard2021padim,liang2025tocoad}; and iv) \textit{Memory-based methods}, which store normal features and detect anomalies based on deviations during inference~\cite{patchcore,bae2023pni}. However, this paradigm faces inherent scalability challenges when extended to multiple classes and prevents cross-category knowledge transfer.

\textbf{One-for-All.} Recent studies~\cite{uniad,guo2023recontrast,guo2025one} have shifted towards training a unified model that can handle multiple categories simultaneously. For example, transformer-based methods (\eg, UniAD~\cite{uniad}, IUF~\cite{iuf}, Dinomaly~\cite{dinomaly}) leverage attention mechanisms to capture inter-category relationships. Data generation approaches, including OmniAL~\cite{omnial} and DiAD~\cite{diad}, synthesize diverse anomaly patterns, while density-based methods, such as CRAD~\cite{crad} and HGAD~\cite{hgad}, learn distributional representations shared across categories. 
However, multi-class models often underperform compared to one-for-one counterparts, and existing evaluations remain limited on relatively small-scale benchmarks ($<$ 40 classes), leaving open questions regarding scalability, cross-domain knowledge transfer, and the feasibility of foundation models.

\textbf{Few-Shot.} 
Vision-language models have enabled new paradigms for anomaly detection with limited training samples. 
WinCLIP~\cite{winclip} and AnomalyCLIP~\cite{anomalyclip} adapt CLIP~\cite{clip} for zero-shot detection through designed text prompts, while PromptAD~\cite{promptad} introduces learnable visual prompts for task adaptation. Recent advances further explore multimodal fusion and alignment~\cite{liu2025continual, liu2025principled}. For example, AA-CLIP~\cite{aaclip} proposes anomaly-aware adaptation mechanisms, FE-CLIP~\cite{feclip} incorporates frequency-domain information, UniVAD~\cite{univad} presents a training-free approach with Contextual Component Clustering for cross-domain detection, and ReMP-AD~\cite{rempad} addresses intra-class variation through retrieval-based token selection and Vision-Language Prior Fusion. 
While these methods demonstrate strong data efficiency, their performance and robustness under domain shifts and generalization across diverse application contexts remain limited.

\section{ADNet}
\label{sec:dataset}
The motivation behind ADNet is to establish a large-scale anomaly detection dataset that spans broader real-world scenarios and can be easily scaled through a standardized data processing pipeline.
We first introduce the data sources and standardization pipeline (Sec.~\ref{sec:datasource}). We then present the data distribution and structure, including both images and textual annotations (Sec.~\ref{sec:datastatistic}). Finally, we discuss the potential challenges associated with using ADNet (Sec.~\ref{sec:challenge}).

\begin{figure}[t]
\centering
\includegraphics[width=\linewidth]{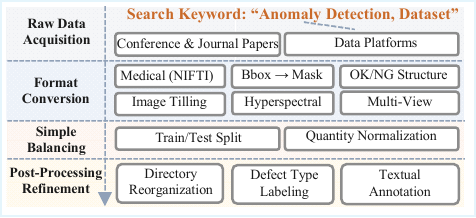}
\caption{\textbf{ADNet construction pipeline.} We collect datasets from diverse sources, standardize formats, balance category-wise samples, and refine annotations with fine-grained defect type labels.}
\label{fig:pipeline}
\end{figure}

\subsection{Data Sources and Standardization Pipeline}
\label{sec:datasource}
We construct ADNet following a standardized pipeline that aggregates 49 publicly available datasets into a unified benchmark, as illustrated in Figure~\ref{fig:pipeline}.
\vspace{-13pt}
\paragraph{Data Sources.}
Existing benchmarks are often constructed from isolated domains or constrained scenarios (\eg, single-device setups), limiting their applicability across diverse real-world settings. To ensure comprehensive domain diversity and mitigate single-source bias, we systematically collect publicly available datasets from three types of sources: academic publications (\eg, conference papers), public data platforms (\eg, Kaggle), and research collaborations~\footnote{The complete list with sources and licenses is provided in the \textit{Supp}.}.
For each candidate source, we apply a set of selection criteria to ensure data quality and usability:
\begin{itemize}[leftmargin=*, nosep]
\item \textit{Data Completeness.} Each dataset must contain normal samples ($>50$ images), anomalous samples ($>20$ images), and corresponding pixel-level masks indicating anomalous regions.
\item \textit{Background Consistency.} Normal and anomalous samples should be acquired under consistent imaging conditions to avoid shortcut learning~\cite{geirhos2020shortcut}, ensuring that models focus on anomalies rather than background variation.
\item \textit{Category Labels.} Each sample must be clearly associated with an explicit category (\eg, hazelnut) to enable standardized organization and downstream usage.
\item \textit{Accessibility.} The dataset must be publicly available with clear licensing for academic use.
\end{itemize}
By adhering to these criteria, we gather 49 datasets that cover a broad spectrum of AD scenarios. To better organize them, we categorize all images into five application domains based on their semantic attributes: Electronics, Industry, Agrifood, Infrastructure, and Medical. More importantly, the principled selection process facilitates scalable and continual expansion of the dataset in the future.

\vspace{-13pt}
\paragraph{Standardization Pipeline.}
To ensure fair comparison and compatibility with existing methods, we design an automated standardization pipeline that converts and unifies all source datasets into the MVTec-style format.
\begin{itemize}[leftmargin=*, nosep]
\item\textit{Format Conversion.} We implement specialized converters to handle heterogeneous data formats, including tiling large-size images (\eg, AITEX textile fabrics, hyperspectral remote sensing) into fixed-size images ($256\times256$ pixels), converting medical imaging formats (NIFTI/DICOM) into slice-level RGB images, and extracting single-view images from multi-view images (\eg, MANTA). 
\item\textit{Sample Balancing.}
To prevent domain biases and ensure computational tractability, we first adopt a 9:1 train/test split for normal samples, then normalize per-category sample quantities with capping limits (a maximum of 500 training samples and 100 normal test samples after splitting, plus up to 100 anomaly samples per defect type).
\end{itemize}
This unified format ensures consistent preprocessing while preserving domain-specific semantics, resulting in balanced categories and reliable statistical evaluation~\footnote{Scripts are provided to support community contributions and extension with new datasets.}.

\vspace{-13pt}
\paragraph{Text Annotation.}
We follow previous studies~\cite{alayrac2022flamingo,apt1988towards,manta} and provide textual descriptions through structured text-visual correspondences. The annotations consist of two components: \textit{spatial localization} and \textit{semantic attributes}.
For spatial localization, each image is divided into a $3\times3$ grid, yielding nine regions. Each defect is assigned a positional descriptor (\eg, ``top-left'', ``middle-center'') to support natural language-based spatial queries.
For semantic attributes, we employ a Large Language Model (\ie, GPT-4o-mini) with domain-specific prompts to generate five visual properties for each anomaly: color, area size, shape, quantity, and reason. Combined with location, all six annotations undergo automated validation followed by manual verification to ensure consistency and accuracy.

\newcommand{\icon}[1]{%
    \includegraphics[width=0.020\linewidth]{#1}%
}

\begin{table}[t]
\centering
\caption{Comparison with existing benchmarks.}
\label{tab:dataset_comparison}
\resizebox{\columnwidth}{!}{%
\begin{tabular}{lclc}
\toprule
\textbf{Dataset} & \textbf{\#Cats} & \textbf{\#Images (Train/Test)} & \textbf{\#Anomalies} \\
\midrule
MVTec-AD~\cite{mvtec}       & 15 & 5,354 (3,629/1,725) & 1,258 \\
VisA~\cite{visa}            & 12 & 10,821 (8,659/2,162) & 1,200 \\
Real-IAD~\cite{realiad}     & 30 & 151,050 (36,465/114,585) & 51,329 \\
Manta-tiny~\cite{manta}     & 38 & 66,068 (21,483/44,585) & 11,025 \\
MulSen-AD~\cite{mulsenad}   & 15 & 2,035 (1,391/644) & 491 \\
Uni-Medical~\cite{unimedical} & 3 & 20,352 (13,339/7,013) & 4,499 \\
MIAD~\cite{miad}            & 7  & 105,000 (70,000/35,000) & 17,500 \\
\midrule
\textbf{ADNet (Ours)} & \textbf{380} & \textbf{196,294 (116,192/80,102)} & \textbf{60,311} \\
\bottomrule
\end{tabular}%
}
\end{table}

\subsection{Dataset Statistics}
\label{sec:datastatistic}

ADNet encompasses 380 categories derived from 49 public datasets. Figure~\ref{fig:dataset_overview} illustrates the hierarchical integration of these source datasets into five target domains: Electronics (77 categories), Industry (154 categories), Agrifood (69 categories), Infrastructure (52 categories), and Medical (28 categories).

\vspace{-13pt}
\paragraph{Data Scale.}
The dataset comprises 196,294 images across 380 categories, including 116,192 training samples and 80,102 test images, among which 19,791 are anomaly-free and 60,311 contain anomalies. 
On average, each category contains 516 images, providing sufficient data while maintaining computational efficiency. 
As shown in Table~\ref{tab:dataset_comparison}, ADNet offers not only a significantly larger data volume, about $46\times$ more than MVTec-AD, but also substantially greater diversity, with a $25\times$ increase in category count over MVTec-AD (15 categories) and a 10$\times$ increase over Manta (38 categories). This reflects broader coverage of real-world scenarios and richer visual variations.

\begin{figure}[t]
\centering
\includegraphics[width=0.95\linewidth]{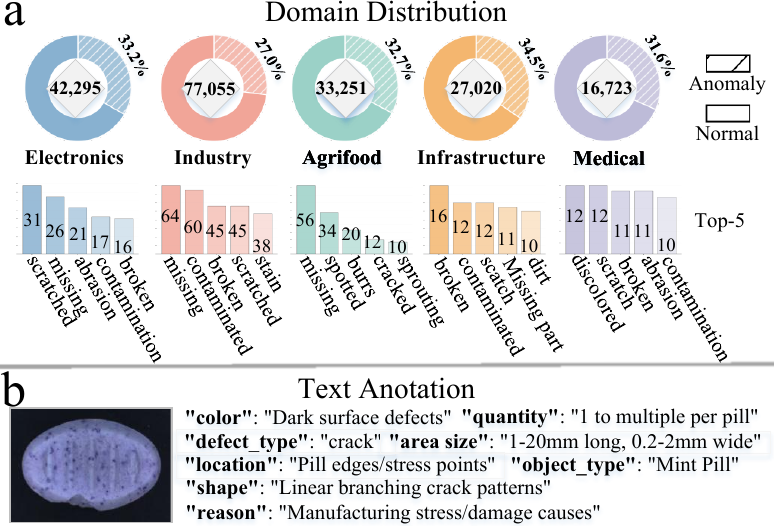}
\caption{\textbf{ADNet domain summary and text annotations.} \textbf{(a)} Total image counts of the five domains with normal vs.\ anomaly proportions. \textbf{(b)} Top-5 defect types and a text-annotation example (image, mask, and JSON fields).}
\label{fig:textanddomain}
\end{figure}

\vspace{-13pt}
\paragraph{Text Information.}
ADNet provides comprehensive textual descriptions, with all annotations stored in a unified JSON format following consistent schemas across diverse industrial domains. The dataset includes fine-grained defect type annotations covering 420 unique defect types across 60,311 anomalous samples. 
Unlike prior datasets that typically use only generic ``defect'' labels, ADNet captures detailed defect semantics. The most prevalent types include \textit{scratch} (2,393 samples, 83 categories), \textit{fragmentation} (2,373 samples, 31 categories), and \textit{missing} (2,227 samples, 30 categories). This dual-level scale (380 categories and 420 defect types) offers finer granularity for evaluation.  

Specifically, each textual description characterizes an anomaly through six attributes: \textit{Location}, indicating the approximate position within the image (\eg, ``top-left''),  \textit{Color}, describing the most salient visual appearance (\eg, ``copper-colored residue''), \textit{Shape}, outlining the geometry of the anomalous region (\eg, ``semicircular notches''), \textit{Area Size}, capturing the relative or physical size (\eg, ``0.2--10 mm linear breaks''), \textit{Quantity}, denoting the number of defect instances (\eg, ``2''), and \textit{Reason}, describing the underlying cause (\eg, ``over-etching'' or ``incomplete separation''). Representative examples are shown in Figure~\ref{fig:textanddomain}.

\subsection{Challenges}
\label{sec:challenge}

\begin{figure}[t]
\centering
\includegraphics[width=\linewidth]{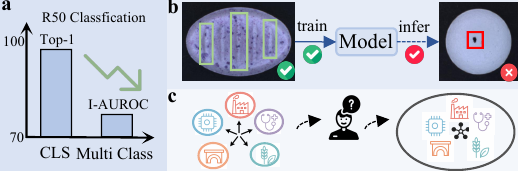}
\caption{Illustration of the key challenges introduced by ADNet. \textbf{a)} Category Catastrophe: existing methods exhibit significant performance degradation on ADNet. \textbf{b)} Context-Dependent Anomalies: ``spots'' are good on mint tablets but bad contamination in a pill. c) Can Anomalies Transfer?, ADNet spans multiple domains, requiring models to learn transferable feature representations.}
\label{fig:challenges}
\vspace{-15pt}
\end{figure}

ADNet introduces several fundamental challenges for existing AD methods, highlighting limitations in generalization, robustness, and adaptability across diverse domains, as shown in Figure~\ref{fig:challenges}.

\vspace{-13pt}
\paragraph{Category Catastrophe.}
Unlike general image classification datasets (\eg, ImageNet), which aim to recognize semantic content, AD datasets focus on identifying deviations from normal samples. To highlight this difference, we conducted a toy experiment by training a ResNet-50 classifier on all 380 categories in ADNet. Normal and anomalous samples were merged within each category, and the data were split into 8:2 train/test ratio. The model achieved 97.9\% top-1 accuracy after 30 epochs. However, when we evaluated advanced multi-class AD methods, their results were notably lower, with I-AUROC scores below 80\%. 

These results indicate that category separation in ADNet is sufficient with minimal inter-class similarity. In contrast, advanced multi-class AD methods, mainly following the reconstruction-based paradigm, show moderate performance, likely due to class confusion~\cite{fan2024revitalizing,uniad}. This evidence suggests that the main challenges in ADNet stem from the intrinsic complexity of anomaly detection rather than from data quality or class-level ambiguity.

\vspace{-13pt}
\paragraph{Context-Dependent Anomalies.}
Given ADNet’s scale of 380 categories, models must maintain robust performance across a wide range of scenarios.
Unlike one-to-one AD approaches, where each model specializes in a single category, multi-class AD methods must detect anomalies across hundreds of diverse categories simultaneously. In many cases, identical visual patterns may carry opposite semantic meanings depending on the context.

For example, ``surface cracks'' are natural textures in baked cookies (Agrifood) but represent critical defects in printed circuit boards (Electronics). Similarly, ``decorative spots'' are expected on mint tablets (Agrifood) yet indicate contamination in pharmaceutical pills (Medical). Therefore, models evaluated on ADNet must acquire context-aware anomaly understanding rather than relying on universal pattern matching. This scalability requirement imposes a rigorous test on whether existing methods can generalize to large-scale, real-world deployment scenarios.

\begin{figure}[t]
\centering
\includegraphics[width=\linewidth]{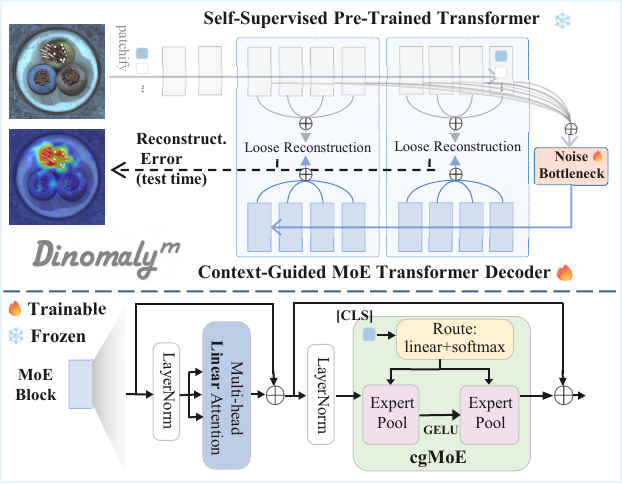}
\caption{\textbf{Overview of Dinomaly$^m$}. We extend Dinomaly by augmenting the decoder with context-guided MoE modules.}
\label{fig:method}
\end{figure}

\vspace{-13pt}
\paragraph{Can Anomalies Transfer?}
The rich textual information and extensive category coverage of ADNet support a range of downstream applications. A central question arises: \textit{Can we build foundation models for anomaly detection that generalize across diverse categories and contexts simultaneously}? In particular, cross-domain transfer, such as whether models trained on industrial defects can generalize to medical or agrifood anomalies, remains largely unexplored. These questions are fundamental to real-world deployment but cannot be adequately addressed using existing single-domain or small-scale benchmarks. ADNet provides the necessary scale and diversity to investigate this direction.

\begin{table*}[t]
\centering
\caption{Results of Multi-class methods on ADNet. We report image-level (I-) AUROC, pixel-level (P-) AUROC, and pixel-level (P-AP) Average Precision (\%) for each domain and the overall average. 
}
\label{tab:multiclass_main}
\resizebox{0.95\textwidth}{!}{%
\begin{tabular}{l|ccc|ccc|ccc|ccc|ccc||ccc}
\toprule
\multirow{2}{*}{\textbf{Method}} & \multicolumn{3}{c|}{\textbf{Electronics}} & \multicolumn{3}{c|}{\textbf{Industry}} & \multicolumn{3}{c|}{\textbf{Agrifood}} & \multicolumn{3}{c|}{\textbf{Infrastructure}} & \multicolumn{3}{c||}{\textbf{Medical}} & \multicolumn{3}{c}{\textbf{Overall}} \\
\cmidrule{2-19}
& \textbf{I} & \textbf{P} & \textbf{P-AP} & \textbf{I} & \textbf{P} & \textbf{P-AP} & \textbf{I} & \textbf{P} & \textbf{P-AP} & \textbf{I} & \textbf{P} & \textbf{P-AP} & \textbf{I} & \textbf{P} & \textbf{P-AP} & \textbf{I} & \textbf{P} & \textbf{P-AP} \\
\midrule
PatchCore~\cite{patchcore}  & 77.5 & 89.0 & \underline{25.2} & 80.4 & 90.4 & 25.8 & \textbf{65.6} & 85.0 & \underline{15.8} & \underline{75.7} & 83.5 & \underline{15.9} & 76.0 & 84.6 & 28.6 & 76.2 & 87.7 & 22.7 \\
UniAD~\cite{uniad}          & 59.9 & 86.2 & 8.2 & 66.1 & 87.5 & 14.4 & 58.6 & \underline{86.1} & 9.1 & 64.4 & 87.1 & 10.8 & 68.2 & 83.7 & 17.1 & 63.4 & 86.7 & 11.8 \\
RD~\cite{rd}                & 68.3 & \underline{90.4} & 14.7 & 70.6 & 89.6 & 19.3 & 59.4 & 82.1 & 9.5 & 63.9 & 83.8 & 9.9 & 67.9 & 87.6 & 26.6 & 66.9 & 87.4 & 15.7 \\
SimpleNet~\cite{simplenet}  & 51.9 & 63.8 & 2.4 & 53.6 & 65.3 & 4.3 & 51.5 & 66.9 & 4.4 & 53.5 & 63.4 & 3.3 & 51.7 & 65.2 & 6.2 & 52.7 & 65.0 & 3.9 \\
ViTAD~\cite{vitad}          & 77.1 & 91.7 & 18.3 & 78.0 & 91.2 & 24.4 & 63.0 & 82.6 & 11.9 & 63.3 & 80.9 & 12.2 & 71.6 & 89.6 & 31.7 & 72.4 & 88.1 & 19.6 \\
DeSTSeg~\cite{destseg}      & 72.6 & 86.1 & 15.3 & 66.1 & 83.3 & 17.9 & 58.0 & 79.7 & 11.1 & 56.4 & 72.1 & 6.8 & 59.1 & 79.0 & 14.0 & 63.9 & 81.0 & 14.0 \\
MambaAD~\cite{mambaad}      & 74.0 & 91.8 & 20.7 & 77.3 & 91.0 & 25.7 & 64.0 & 83.1 & 12.0 & 69.4 & 85.8 & 13.4 & 72.5 & 89.4 & 29.9 & 72.7 & 88.8 & 20.7 \\
Dinomaly~\cite{dinomaly}    & \underline{82.7} & \underline{92.8} & 23.3 & \underline{85.2} & \underline{93.8} & \underline{30.3} & 63.4 & 85.7 & 13.3 & 73.0 & 87.0 & 15.4 & \underline{77.6} & \underline{90.3} & \underline{32.1} & \underline{78.5} & \underline{91.0} & \underline{23.9} \\
LGC-AD~\cite{fan2024revitalizing}     & 73.7 & 92.4 & 19.0 & 74.8 & 91.8 & 23.6 & 63.7 & 83.2 & 11.3 & 69.3 & \underline{87.4} & 13.0 & 74.0 & 89.6 & 31.8 & 71.8 & 89.6 & 19.6 \\
\midrule
\textit{Dinomaly$^m$}     & \textbf{88.9} & \textbf{95.6} & \textbf{32.6} & \textbf{90.6} & \textbf{95.8} & \textbf{38.8} & \underline{65.3} & \textbf{88.4} & \textbf{16.2} & \textbf{78.8} & \textbf{88.9} & \textbf{21.6} & \textbf{81.4} & \textbf{91.6} & \textbf{36.0} & \textbf{83.2} & \textbf{93.1} & \textbf{30.6} \\
\bottomrule
\end{tabular}%
}
\vspace{-2mm}
\end{table*}

\section{Baseline}
\label{sec:method}
To validate and explore the potential of ADNet, we introduce a baseline model, \textbf{Dinomaly$^m$}, a context-guided extension of Dinomaly~\cite{dinomaly} based on a mixture-of-experts framework. The vanilla Dinomaly employs a frozen DINOv2~\cite{dinov2} encoder, a noisy bottleneck, and a lightweight Transformer decoder to reconstruct clean features. Only the bottleneck and decoder are trained, and anomalies are detected via reconstruction errors. 
While effective on existing benchmarks, scaling to ADNet results in only moderate performance (78.5\% I-AUROC). We attribute this degradation to the fixed-capacity decoder, which struggles to model the diverse visual patterns across hundreds of categories and to the strong context dependence of anomaly semantics.

We augment Dinomaly by introducing a context-guided mixture-of-experts (cgMoE) module into the decoder, enhancing its representational capacity with minimal runtime overhead and enabling context-specific reconstruction. As illustrated in Figure~\ref{fig:method}, given an input image $I$, the frozen encoder extracts features $f_{E}$ and produces a final-layer [CLS] token $\mathbf{z}_{\text{cls}} \in \mathbb{R}^d$ that captures the global context. In each decoder block, a gating network uses $\mathbf{z}_{\text{cls}}$ to compute soft routing weights over $K$ experts:
\begin{equation}
\mathbf{g} = [g_1, \ldots, g_K] = \text{softmax}(\mathbf{W}_g \mathbf{z}_{\text{cls}}) \in \mathbb{R}^K,
\end{equation}
where $\mathbf{W}_g \in \mathbb{R}^{K \times d}$ is a learnable parameter matrix, and $g_k \in \mathbb{R}$ is the scalar weight for expert $k$. 
Each expert $k$ is a two-layer feed-forward network with parameters $\mathbf{W}_k^{(1)} \in \mathbb{R}^{h \times d}$ and $\mathbf{W}_k^{(2)} \in \mathbb{R}^{d \times h}$, where $h$ is the hidden dimension. These are adaptively mixed based on the routing weights:

\begin{equation}
\text{cgMoE}(\mathbf{x}) = 
\sum_{k=1}^{K} g_k \mathbf{W}_k^{(2)}\text{GELU}(\sum_{k=1}^{K} g_k \mathbf{W}_k^{(1)}\mathbf{x} ).
\end{equation}
All decoder blocks share the same $\mathbf{z}_{\text{cls}}$ to ensure consistent routing, and the decoder outputs layer-wise features $f_D$ for reconstruction. Training is performed using a reconstruction loss defined as:
\begin{equation}
\mathcal{L}_{\textrm{total}} = \mathcal{D}_{\text{cos}}(\mathcal{F}(f_E), \mathcal{F}(sg(f_D))),
\end{equation}
where $\mathcal{D}_{\text{cos}}$ is cosine distance, $\mathcal{F}(\cdot)$ flattens the feature maps, and $sg(\cdot)$ denotes the hard-mining mechanism that shrinks gradients of well-restored feature points~\cite{dinomaly}. During training, the encoder is frozen, while the bottleneck, expert weights, and gating network are optimized jointly. At inference, anomaly scores are computed from reconstruction errors, with routing performed once per image.

\section{Experiments}
\subsection{Experimental Setup}
\noindent\textbf{Evaluation Settings.}
We evaluate methods under two settings: \textit{(1) Single-Class Evaluation:} Following the standard protocol~\cite{patchcore}, we train 380 independent category-specific models, one per category. This setting provides an upper-bound performance. \textit{(2) Multi-Class Evaluation:} We train a single unified model on all 380 categories and evaluate it across all test sets. This setting directly assesses a model's ability to handle diverse contexts.

\noindent\textbf{Metrics and Implementation Details.}
We report standard image-level (I-) and pixel-level (P-) Area Under the Receiver Operating Characteristic curve (AUROC) scores~\cite{patchcore,mvtec} to evaluate model performance in anomaly detection and localization, respectively. 
We implemented baseline methods using the ADer toolbox~\cite{ader}.
For multi-class evaluation, we trained models for 30 epochs following the default setups in ADer. Dinomaly variants were trained for 100k iterations (27.5 epochs) with ViT-B and $280^2$ input. Complete details, additional metrics, and comprehensive experimental results are provided in the \textit{Supp}.

\subsection{Multi-Class Results}
\label{sec:multiclass_results}

We evaluated multi-class performance on ADNet by comparing our Dinomaly$^m$ against nine representative baselines across three major paradigms: memory-based (PatchCore), reconstruction-based (UniAD, RD, ViTAD, MambaAD, Dinomaly, LGC-AD), and synthesis-based methods (SimpleNet, DeSTSeg). These baselines consistently achieve over 95\% I-AUROC on MVTec-AD~\cite{mvtec}, making them strong candidates for assessing the challenges posed by ADNet. Results are shown in Table~\ref{tab:multiclass_main}.

Dinomaly$^m$ achieved an overall 83.2\% I-AUROC, outperforming the baseline (Dinomaly at 78.5\%) by a margin of 4.7\% I-AUROC. This improvement validates the effectiveness of incorporating context-adaptive capacity. The mixture-of-experts decoder enables specialized sub-networks for different semantic contexts, directly addressing context-dependent anomaly semantics without incurring additional inference costs. This architectural choice proves particularly effective for ADNet's multi-domain.

However, this performance remains far below the 95-99\% typically reported on MVTec-AD, highlighting the substantial scalability challenges introduced by ADNet. Although P-AUROC remains relatively high (93.1\%), pixel-level Average Precision (P-AP) remains considerably low across all methods (3.9\%-30.6\%). This gap reveals that establishing precise anomaly boundaries under diverse semantic contexts remains a fundamental challenge for current AD approaches.

\begin{figure}[t]
\centering
\includegraphics[width=\columnwidth]{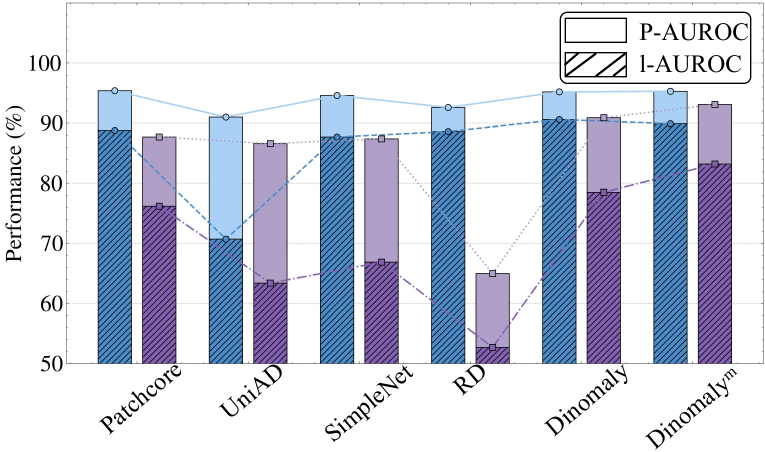}
\caption{Comparison between single-class and multi-class settings. We report results averaged across all 380 categories.
}
\label{fig:single_vs_multi}
\vspace{-2mm}
\end{figure}

\noindent\textbf{Single-Class \textit{vs.} Multi-Class.}

\label{sec:single_vs_multi}
To quantify the gap between single-class models and multi-class models, we conducted five representative methods under both settings. Figure~\ref{fig:single_vs_multi} illustrates the stark contrast between single-class (average 380 models) and training a single model on all categories. In the single-class setting, all methods achieved 87-90\% I-AUROC, demonstrating that individual ADNet categories are inherently learnable. This performance confirms that ADNet's challenge stems not from ambiguity within categories but from the complexity of unified multi-category modeling. Under the multi-class setting, however, we observe substantial and method-dependent performance degradation. SimpleNet collapsed with a 35.9\% absolute drop in I-AUROC, and RD degraded by 20.8\%. In contrast, PatchCore and Dinomaly exhibited more moderate degradation (approximately 12\% each).

These degradation patterns reveal that architectural design is critical for multi-class robustness. Methods employing context-invariant strategies (SimpleNet's fixed augmentation templates, RD's single student-teacher architecture) collapse under context diversity. Conversely, incorporating implicit context-awareness (PatchCore's memory-based matching) degrades more gracefully. These observations motivate our Dinomaly$^m$'s explicit context-adaptive design through mixture-of-experts, which enables specialized sub-networks to handle diverse semantic contexts effectively.

\begin{table}[t]
\small
\centering
\caption{Zero-shot cross-domain transfer. Models trained solely on Industry are tested on other domains. Numbers show performance drop from in-domain training.}
\vspace{-5pt}
\label{tab:cross_domain}
\resizebox{0.5\textwidth}{!}{%
\begin{tabular}{l|c|c|c|c|c|c}
\toprule
\multirow{2}{*}{\textbf{Method}} & \textbf{In-Domain} & \multicolumn{4}{c|}{\textbf{Cross-Domain (Zero-Shot Transfer)}} & \textbf{Avg Transfer} \\
\cmidrule{2-7}
& \textbf{Industry} & \textbf{Electronics} & \textbf{Agrifood} & \textbf{Infrastructure} & \textbf{Medical} & \textbf{Gap} \\
\midrule
PatchCore   & 81.8 & 48.8 \gd{-33.0} & 55.6 \gd{-26.2} & 60.1 \gd{-21.7} & 58.4 \gd{-23.4} & \textcolor{red}{-26.1\%} \\
UniAD       & 73.4 & 48.3 \gd{-25.1} & 56.1 \gd{-17.3} & 58.7 \gd{-14.7} & 64.1 \gd{-9.3} & \textcolor{red}{-16.6\%} \\
RD          & 71.9 & 55.3 \gd{-16.6} & 57.0 \gd{-14.9} & 57.8 \gd{-14.1} & 61.1 \gd{-10.8} & \textcolor{red}{-14.1\%} \\
VITAD       & 82.2 & 58.5 \gd{-23.7} & 55.5 \gd{-26.7} & 52.8 \gd{-29.4} & 58.8 \gd{-23.4} & \textcolor{red}{-25.8\%} \\
Dinomaly    & 90.1 & 54.3 \gd{-35.8} & 55.2 \gd{-34.9} & 59.0 \gd{-31.1} & 55.5 \gd{-34.6} & \textcolor{red}{-34.1\%} \\
\bottomrule
\end{tabular}%
}
\vspace{-5pt}
\end{table}

\begin{table}[t]
\centering
\caption{Zero-shot and few-shot anomaly detection performance on ADNet. We report I-AUROC and P-AUROC (\%) across different shot settings.}
\label{tab:few_shot}
\resizebox{0.8\columnwidth}{!}{
\begin{tabular}{l|cc|cc|cc}
\toprule
\multirow{2}{*}{Method} & \multicolumn{2}{c|}{0-shot} & \multicolumn{2}{c|}{1-shot} & \multicolumn{2}{c}{4-shot} \\
\cmidrule{2-7}
& I & P & I & P & I & P \\
\midrule
WinCLIP~\cite{winclip} & 60.3 & 74.7 & 63.9 & 85.9 & 66.4 & 86.5 \\
\midrule
Dinomaly & - & - & 68.6 & 89.9 & 72.6 & 90.6 \\
Dinomaly$^m$ & - & - & 66.8 & 88.6 & 72.3 & 91.1 \\
\bottomrule
\end{tabular}
}
\vspace{-5pt}
\end{table}

\subsection{Cross-Domain Generalization}

\label{sec:cross_domain}

Real-world deployment frequently demands that models generalize to new application domains with limited or no domain-specific training data. To evaluate zero-shot cross-domain transferability, we train models exclusively on the Industry domain (154 categories) and evaluate their performance on the remaining four domains without any fine-tuning, as shown in Table~\ref{tab:cross_domain}.

All methods experience substantial performance degradation under zero-shot cross-domain evaluation, with average drops ranging from 14.1\% (RD) to 34.1\% (Dinomaly). Notably, high in-domain performance does not guarantee cross-domain transferability: Dinomaly achieves the highest in-domain performance (90.1\% on Industry) yet exhibits the most severe transfer gap, whereas RD with moderate source domain performance (71.9\%) transfers most robustly across domains.

\subsection{Few-shot Multi-Class Anomaly Detection}

We evaluated the few-shot capability of multi-class models (Dinomaly and Dinomaly$^m$) under 1-shot and 4-shot settings, with WinCLIP~\cite{winclip} serving as the single-class few-shot baseline (Table~\ref{tab:few_shot}). 
Both multi-class models demonstrated strong sample efficiency. Dinomaly yielded 68.6\% and 72.6\% I-AUROC with 1-shot and 4-shot, while Dinomaly$^m$ exhibited similar results of 66.8\% and 72.3\%.
More remarkably, at the 4-shot level, Dinomaly and Dinomaly$^m$ not only surpassed one-for-one few-shot methods (WinCLIP) but also outperformed several full-shot multi-class baselines, including UniAD (63.4\%), RD (66.9\%), and ViTAD (72.4\%, as shown in Table~\ref{tab:multiclass_main}).

\subsection{Performance Across Dataset Scales}
\label{sec:scale_saturation}
To systematically understand how methods scale with increasing category diversity, we evaluated performance on ADNet subsets of progressively increasing sizes: 50, 100, 200, and 380 categories, as shown in Table~\ref{tab:scale_saturation}. To isolate the degeneration from varied categories, we also evaluated a fixed 50-category subset across different training scales.

Most methods exhibited non-linear performance degradation characterized by a rapid decline from 50 to 200 categories, followed by a plateau phase. For instance, RD experienced a sharp 9.0\% drop in 200 categories but only an additional 0.8\% decline when scaling to 380 categories. This pattern suggests that methods encounter a critical complexity threshold around 100-200 categories, beyond which additional categories introduce diminishing marginal difficulty.
Our Dinomaly$^m$ achieved the optimal results, exhibiting -6.0\% degradation on the fixed subset ADNet-50. This result demonstrates that MoE-based context-adaptive capacity effectively mitigates negative transfer at scale.

\begin{table}[t]
\centering
\caption{Performance degradation as dataset scale increases. All methods use multi-class evaluation on ADNet subsets of varying sizes. For ADNet-100/200/380, we also report performance on the fixed ADNet-50 subset to demonstrate negative transfer effects.}
\vspace{-5pt}
\label{tab:scale_saturation}
\resizebox{\columnwidth}{!}{%
\begin{tabular}{l|cc|cccc|cccc|cccc}
\toprule
\multirow{3}{*}{\textbf{Method}} & \multicolumn{2}{c|}{\textbf{ADNet-50}} & \multicolumn{4}{c|}{\textbf{ADNet-100}} & \multicolumn{4}{c|}{\textbf{ADNet-200}} & \multicolumn{4}{c}{\textbf{ADNet-380}} \\
\cmidrule{2-15}
& \multicolumn{2}{c|}{All} & \multicolumn{2}{c}{All} & \multicolumn{2}{c|}{ADNet-50} & \multicolumn{2}{c}{All} & \multicolumn{2}{c|}{ADNet-50} & \multicolumn{2}{c}{All} & \multicolumn{2}{c}{ADNet-50} \\
\cmidrule{2-15}
& \textbf{I} & \textbf{P} & \textbf{I} & \textbf{P} & \textbf{I} & \textbf{P} & \textbf{I} & \textbf{P} & \textbf{I} & \textbf{P} & \textbf{I} & \textbf{P} & \textbf{I} & \textbf{P} \\
\midrule
UniAD       & 68.5 & 88.5 & 67.9 & 88.6 & 67.2 & 87.6 & 67.3 & 88.2 & 65.6 & 86.9 & 63.4 & 86.6 & 61.9 & 85.1 \\
RD          & 76.7 & 89.7 & 73.4 & 89.7 & 73.5 & 88.5 & 67.7 & 88.0 & 67.8 & 87.3 & 66.9 & 87.4 & 68.6 & 86.9 \\
SimpleNet   & 56.9 & 76.5 & 54.3 & 77.5 & 56.5 & 77.7 & 52.6 & 67.9 & 52.6 & 67.3 & 52.7 & 65.0 & 53.2 & 65.4 \\
Dinomaly    & 86.0 & 93.7 & 83.2 & 93.2 & 82.8 & 92.4 & 82.3 & 92.4 & 81.1 & 91.8 & 78.5 & 91.0 & 78.4 & 90.9 \\
\midrule
Dinomaly$^m$ & 89.0 & 94.9 & 86.7 & 94.6 & 86.0 & 93.8 & 85.9 & 94.0 & 84.9 & 93.4 & 83.2 & 93.1 & 83.0 & 92.9 \\
\bottomrule
\end{tabular}%
}
\end{table}

\subsection{Ablation Study}
\label{sec:ablation}
We conducted comprehensive ablation studies on Dinomaly$^m$ using the full ADNet benchmark. Table~\ref{tab:ablation} systematically validates our key design choices.

\noindent\textbf{Impact of Expert Count.}
Our cgMoE replaces standard feed-forward layers (FFN) with a weighted ensemble of expert networks, where routing weights are dynamically computed from the encoder's [CLS] token. Starting from a single FFN baseline (78.5\% I-AUROC), we observed consistent improvements with increasing expert count. Notably, scaling to 12 experts yielded only marginal gains (83.3\%), indicating that 8 experts sufficiently capture ADNet's context diversity without introducing redundant capacity. 

\noindent\textbf{Routing Strategy.}
We compared two routing strategies: using the encoder's [CLS] token versus the [CLS] token in each MoE layer in the decoder. Encoder-based routing (83.2\%) outperformed decoder-based routing (82.7\%), validating the hypothesis that pre-encoded representations contain richer semantic context for expert selection. 

\noindent\textbf{Backbone Scaling.}
We further evaluated Dinomaly$^m$ with ViT-Large encoder. The larger backbone achieved 84.3\% I-AUROC, representing a 1.1\% absolute improvement over ViT-Base (83.2\%). This result demonstrates that cgMoE and backbone scaling provide complementary benefits. However, we adopted ViT-Base as our default configuration, as it offers the optimal accuracy-efficiency trade-off for practical deployment scenarios where computational resources are constrained.

\begin{figure}[t]
\centering
\includegraphics[width=0.95\linewidth]{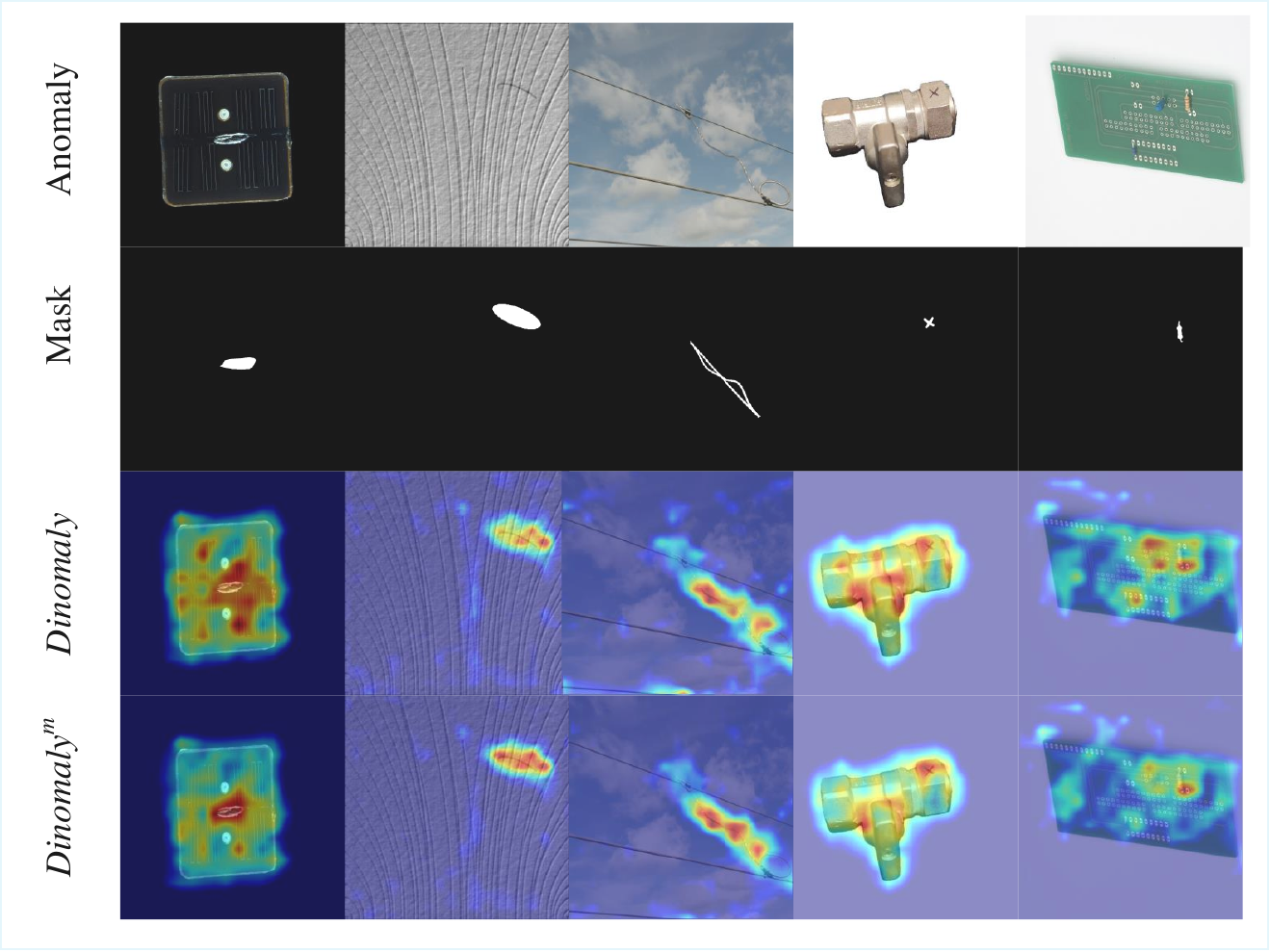}
\vspace{-10pt}%
\caption{Result Visualization of Dinomaly and Dinomaly$^{m}$.
}
\vspace{-10pt}
\label{fig:viza}
\end{figure}

\noindent\textbf{Visualization.} We further visualize anomaly heatmaps by comparing our Dinomaly$^m$ with the baseline Dinomaly, as shown in Figure~\ref{fig:viza}. The results show that Dinomaly$^m$ attends more precisely to anomalous regions, producing clearer and more focused localization.

\begin{table}[t]
\centering
\caption{Ablation of Dinomaly$^m$ design on ADNet.}
\label{tab:ablation}
\resizebox{0.85\columnwidth}{!}{
\begin{tabular}{l|cc}
\toprule
\textbf{Configuration} & \textbf{I-AUROC} & \textbf{P-AUROC} \\
\midrule
Single FFN (baseline) & 78.5 & 89.4 \\
\midrule
cgMoE-2 experts & 79.6 & 91.5 \\
cgMoE-4 experts & 81.6 & 92.1 \\
\textbf{cgMoE-8 experts (ours)} & 83.2 & 93.1 \\
cgMoE-12 experts & 83.3 & 92.8 \\
\hline
cgMoE-8, decoder [CLS] routing & 82.7 & 92.8 \\
cgMoE-8, ViT-Large & 84.3 & 93.6 \\
\bottomrule
\end{tabular}
}
\end{table}

\section{Conclusion}

We introduced ADNet, a large-scale benchmark with 380 categories and 196,294 images spanning five domains, and proposed Dinomaly$^m$, a context-guided mixture-of-experts extension that achieves 83.2\% I-AUROC. Our evaluation reveals that current AD methods struggle with both scalability (78.5\% on 380 categories vs.\ 92–99\% on small benchmarks) and cross-domain transfer (14–34\% degradation), highlighting the substantial challenges posed by ADNet.

\noindent \textbf{Limitation and Outlook.}
Despite this progress, ADNet's scale remains limited, with an imbalanced domain distribution (\textit{e.g.}, Industry with 154 categories vs. Medical with only 28 categories). In future work, we plan to continuously expand ADNet by incorporating more diverse datasets to achieve better domain balance and larger category coverage. We will also explore stronger baseline methods and investigate the feasibility of building foundation models for anomaly detection that can effectively handle the two core challenges: scaling to thousands of categories while maintaining performance, and understanding context-dependent anomalies where identical visual patterns exhibit opposite meanings across different application domains.

\clearpage
{
    \small
    \bibliographystyle{ieeenat_fullname}
    \bibliography{main}
}

\end{document}